\documentclass[runningheads]{llncs}
\usepackage{graphicx}

\usepackage{comment}
\usepackage{amsmath,amssymb} 
\usepackage{color}
\usepackage{subcaption}

\usepackage[accsupp]{axessibility}  

\begin{document}
\pagestyle{headings}
\mainmatter
\def\ECCVSubNumber{008}  

\title{Adversarial Vulnerability of Temporal Feature Networks for Object Detection} 


\titlerunning{Adversarial Vulnerability of Temporal Feature Networks}
%
\author{Svetlana Pavlitskaya\inst{1} \and
Nikolai Polley\inst{2} \and
Michael Weber\inst{1} \and
J. Marius Zöllner\inst{1,2}}
\authorrunning{S. Pavlitskaya et al.}
%
\institute{FZI Research  Center  for  Information  Technology, 76131 Karlsruhe, Germany \\
\email{pavlitskaya@fzi.de} \and
Karlsruhe Institute of Technology (KIT), 76131 Karlsruhe, Germany\\
}
\maketitle

\begin{abstract}
Taking into account information across the temporal domain helps to improve environment perception in autonomous driving. However, it has not been studied so far whether temporally fused neural networks are vulnerable to deliberately generated perturbations, i.e. adversarial attacks, or whether temporal history is an inherent defense against them.  In this work, we study whether temporal feature networks for object detection are vulnerable to universal adversarial attacks. We evaluate attacks of two types: imperceptible noise for the whole image and locally-bound adversarial patch. In both cases, perturbations are generated in a white-box manner using PGD. Our experiments confirm, that attacking even a portion of a temporal input suffices to fool the network. We visually assess generated perturbations to gain insights into the functioning of attacks. To enhance the robustness, we apply adversarial training using 5-PGD. Our experiments on KITTI and nuScenes datasets demonstrate, that a model robustified via K-PGD is able to withstand the studied attacks while keeping the mAP-based performance comparable to that of an unattacked model. 
\end{abstract}

\section{Introduction}

Deep neural networks (DNNs) have become an indispensable component of environment perception in autonomous driving systems. The inherent vulnerability of DNNs to adversarial attacks~\cite{szegedy2013intriguing} makes adversarial robustness one of the crucial requirements before their wide adoption in autonomous vehicles is possible. Recent studies~\cite{brown2017adversarial,eykholt2018robust,lee2019physical} demonstrate that adversarial attacks can be performed in the real world and thus present a significant threat to self-driving cars.

Previous works have already investigated adversarial vulnerability of DNNs for specific tasks like object detection~\cite{lee2019physical} or semantic segmentation~\cite{metzenKBF17,nesti2022evaluating}, and also of DNNs with specific architectures like sensor fusion~\cite{tu2021exploring} or multi-task learning~\cite{maoGNRSYV20}. In this work, we focus on temporal feature networks as a model under attack. Although the emphasis of recent studies was mostly on LiDAR data used in conjunction with camera images, the temporal fusion is a further approach to increase the object detection accuracy, which deserves attention. There has been very little research into fusing multiple images in the temporal dimension for object detection. Object detectors with temporal fusion, however, receive more context data from previous images and outperform single-image object detectors~\cite{weber2021}. In particular, the prediction of obstructed objects that are visible in previous images might be a possible advantage.

\begin{figure}[t]
    \centering
	\includegraphics[width=1.0\textwidth]{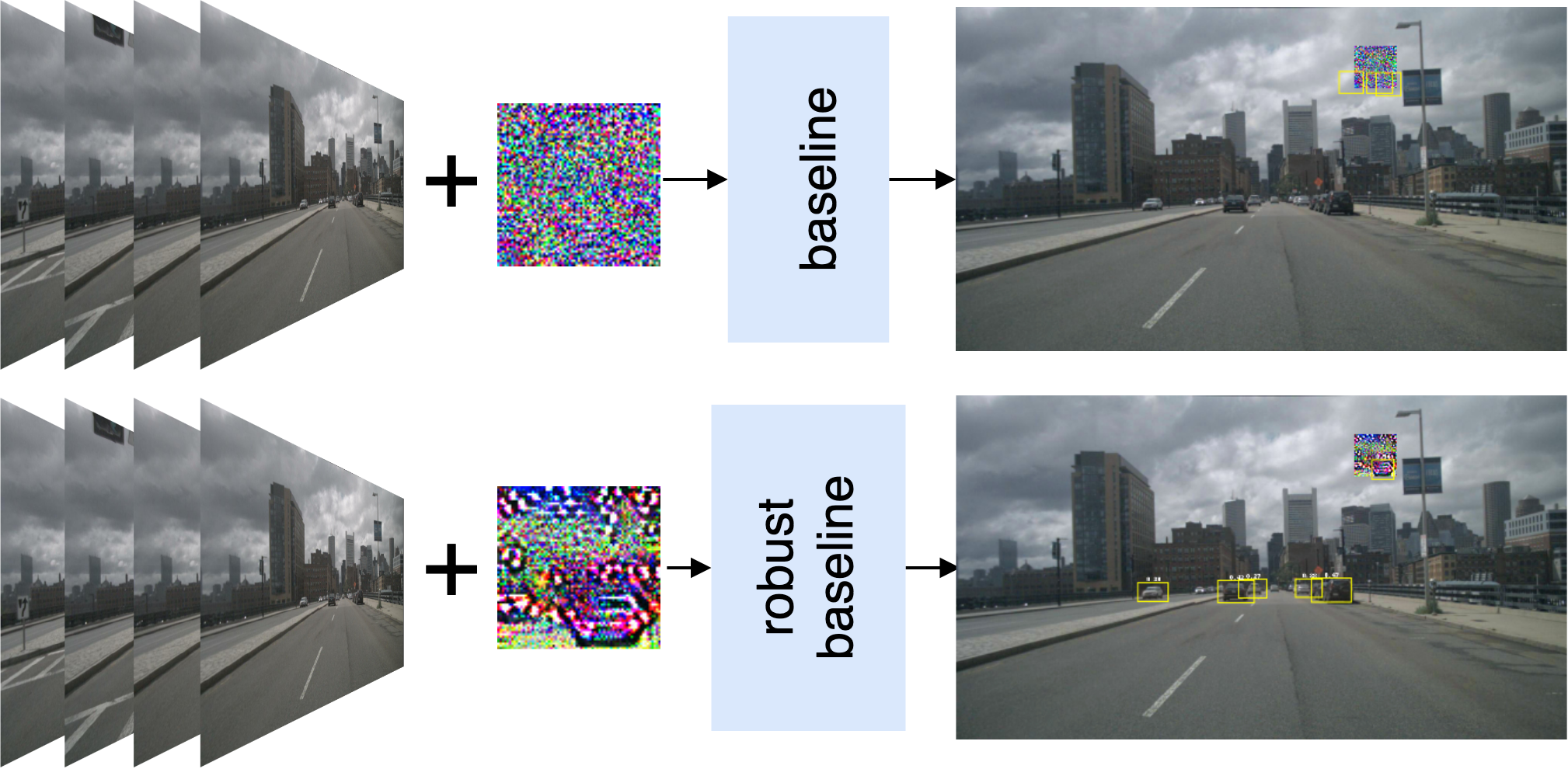}
	\caption{We evaluate universal attacks on temporal feature networks for object detection (here -- with a universal patch). Adversarial patch suppresses all detections of an undefended baseline and creates fake detections instead (top). Detection of ground truth objects by a robust model is no longer restrained by an adversarial patch (bottom). Note, that the patch against robust models has a more complex structure}
	\label{concept_pic}
\end{figure}

To the best of our knowledge, we are the first to explore attacks and defenses for this type of DNNs. As an exemplary model under attack, we consider the late slow fusion architecture for object detection proposed by Weber et al.~\cite{weber2021}, which is in turn inspired by DSOD~\cite{Shen_2017_ICCV}
and expanded to incorporate several input images. The model proposed by  Weber et al. outperformed its non-temporal counterpart, thus demonstrating the importance of the temporal history for the object detection accuracy. The main goal of our work is to study, whether CNNs with temporal fusion are prone to adversarial attacks and what is the impact of the temporal history on the adversarial vulnerability of these models.

For this, we evaluate two variants of universal white-box attacks against this model: with an adversarial patch (see Figure \ref{concept_pic}) and with adversarial noise. In our setting, a single instance of malicious input is generated to attack all possible data. This way, if printed out, the generated patch can be used for an attack in the real world. The latter was already demonstrated for state-of-the-art object detectors in previous works~\cite{brown2017adversarial,songEEF0RTPK18,thysRG19}. To increase the adversarial robustness of the model under attack, we consider adversarial training using K-PGD~\cite{madry2017towards}. 

\section{Related Work}

\subsection{Adversarial Attacks}

Since the discovery of adversarial attacks by Szegedy et al.~\cite{szegedy2013intriguing}, a number of algorithms to attack DNNs have been proposed, the most prominent being the Fast Gradient Sign Method (FGSM) ~\cite{goodfellow2014explaining} and the Projected Gradient Descent (PGD)~\cite{madry2017towards}. 

While the mentioned attacks are traditionally performed in a per-instance manner, i.e. an adversarial perturbation is applied to a single image in order to fool a model, universal perturbations, that are able to attack multiple instances, are also possible~\cite{moosavi2017universal}. Universal adversarial inputs pose a special threat, because they are also able to attack images beyond the training data.  

Another line of research aims at developing physical adversarial attacks. For this, visible adversarial perturbations are generated within a certain image area. The resulting \textit{adversarial patch} can then be printed out to perform an attack in the real world. After their introduction by Brown et al. in~\cite{brown2017adversarial}, adversarial patches have been shown to successfully fool various deep learning models, including object detectors~\cite{lee2019physical,pavlitskaya2022feasibility}, semantic segmentation networks~\cite{nesti2022evaluating} and end-to-end driving models~\cite{pavlitskaya2020feasibility}. A combination of adversarial patches with universal attacks is especially interesting. Taking into consideration the transferability of adversarial examples across DNNs, such attacks might also be performed in a black-box manner in the real world~\cite{thysRG19,xuZ0FSCCWL20}.

Although no previous work on adversarial vulnerability of temporal fusion networks is known, a certain effort was already made in the community to develop adversarial attacks against sensor fusion models~\cite{yu2020investigating,tu2021exploring}. In particular, the work by Yu et al. has revealed, that late fusion is more robust against attacks than early fusion~\cite{yu2020investigating}. The evaluated dataset, however, is relatively small with only 306 training and 132 validation samples from the KITTI dataset.

\subsection{Adversarial Training}
Adversarial training (AT) is currently one of the few defenses that are able to combat even strong attacks~\cite{athalye2018obfuscated}. It consists in training a DNN while adding adversarial inputs to each minibatch of training data. It has recently been shown, that adversarial training not only increases the robustness of neural networks to adversarial attacks but also leads to better interpretability~\cite{tsipras2018robustness}.

While the idea originates from~\cite{goodfellow2014explaining}, the first strong defense was demonstrated with a multi-step PGD algorithm (the K-PGD adversarial training)~\cite{madry2017towards}. For each minibatch, a forward/backward step is first executed $k$ times to generate an adversarial input and then a single forward/backward step follows, which aims to update the model parameters. The PGD loop thus drastically increases the overall training time. For this reason, K-PGD adversarial training is intractable for large datasets. 

One of the recently proposed strategies to speed up adversarial training is the so-called AT for free~\cite{shafahi2019adversarial}, which reuses gradient information during training. Instead of performing separate gradient calculations to generate adversarial examples during training, adversarial perturbations and model parameters are updated simultaneously in a single backward pass. This way, multiple FGSM steps are performed on a single minibatch to simulate the PGD algorithm while concurrently training the model. The authors were the first to apply adversarial training to the large-scale ImageNet classification task. The robustified models demonstrate resistance to attacks, comparable to that of K-PGD, while being 7 to 30 times faster. The approach, however, is still more time-consuming than standard training.

A similar method to accelerate adversarial training named YOPO (You Only Propagate Once) is proposed by Zhang et al.~\cite{zhang2019you}. In YOPO, the gradients of the early network layers are frozen and reused to generate an adversarial input. YOPO is four to five times faster than K-PGD, although the results are only provided for relatively small datasets. YOPO reaches a performance similar to the free adversarial training but is less computationally expensive.

Most recently, a further approach to accelerate adversarial training was proposed by Wong et al.~\cite{wong2020fast}. Starting with the assumption, that iterative attacks like K-PGD do not necessarily lead to more robust defenses, the authors propose to use R-FGSM AT instead. R-FGSM  applies FGSM after a random initialization. This adversarial training method is claimed to be as effective as K-PGD.

Enhancing adversarial training to increase robustness against universal attacks was first addressed by Shafahi et al.~\cite{shafahi2020universal}.  Each training step uses FGSM to update a universal adversarial perturbation, which is then simultaneously used to update the model parameters. The proposed extension to adversarial training introduces almost no additional computational cost, which makes adversarial training on large datasets possible.

A further approach to harden DNNs against universal attacks is the shared adversarial training, proposed in~\cite{mummadi2019defending}. To generate a shared perturbation, each batch is split into heaps, which are then attacked with single perturbations. These perturbations are aggregated and shared across heaps and further used for standard adversarial training.

\section{Adversarial Attacks}

\subsection{Threat Model}

We consider two types of white-box attacks in this work: adversarial patch and adversarial noise. All attacks are performed in a universal manner, i.e. a single perturbation is used to to attack all images~\cite{moosavi2017universal}. 
We use a slightly modified version of the PGD algorithm~\cite{madry2017towards} for the attack. In order to apply PGD in a universal manner, an empty mask is introduced, which is added to each input image. We then only update this mask in contrast to the original PGD, which updates the input images directly.

We use Adam~\cite{kingma2014adam} for faster PGD convergence as suggested in~\cite{carlini2017}. We also do not take a sign of the gradients but use actual gradient values instead. In an unsigned case, pixels, that strongly affect the prediction, can be modified to a much larger extent than less important pixels. Noise initialization strategy is a further setting, influencing training speed. For the FGSM attack, the advantage of initialization with random values has already been shown~\cite{tramer2017ensemble}. We perform patch-based attacks with randomly initialized patch values and noise-based attacks initialized with Xavier~\cite{glorot2010understanding}.

\subsection{Adversarial Training}

Adversarial training expands the training dataset with adversarial examples, created on the fly during training. Training on this dataset should enable the model to predict the correct label even in the presence of an attack.

We consider the established K-PGD AT with patch/noise generated for each input sequence. We create a single adversarial example (with either patch or noise) for all images in an input sequence. Thus, the generated adversarial perturbation is not universal and is only intended to fool the data from the current batch. The usage of this approach is motivated by a recent observation, that defending against non-universal attacks also protects against universal attacks~\cite{mummadi2019defending}.

Although Free~\cite{shafahi2019adversarial} and YOPO~\cite{zhang2019you} approaches offer a considerable speed up in training, they are not applicable in our case. Both algorithms require the same loss function to train a model and create adversarial examples. This is only possible for untargeted attacks, where the goal is to maximize the loss for the correct class. Instead, we want to perform object vanishing attacks, which require different loss functions than training the model. This way, reusing the gradients, already computed for the parameter update step, as foreseen in these AT algorithms, cannot be performed in our case.

\section{Experimental Setup}

\subsection{Dataset}

For the evaluation, we consider the late slow fusion architecture for object detection, proposed by Weber et al. ~\cite{weber2021}. While the original work by Weber et al. has focused on the KITTI Object Detection dataset~\cite{geigerLU12}, we additionally run our experiments on the nuScenes data~\cite{nuscenes2019}. We focus on the models with four input frames with equal temporal distance.

\subsubsection{KITTI}

Images from the object detection benchmark are resized to 1224$\times$370. We only select images, that have the corresponding three temporally preceding frames, delivered in the dataset, resulting in 3689 sequences of length four for training and 3754 sequences of length four for validation.

\subsubsection{nuScenes}
The dataset contains 850 annotated scenes, whereas each scene contains about 40 keyframes per camera, taken at a frequency of 2Hz. We generate input sequences with a length of four from the keyframes as follows: if keyframes \textit{ \{a, b, c, d, e\}} belong to the same scene, the two sequences \textit{ \{a, b, c, d\} } and \textit{ \{b, c, d, e\}} are created. For the training dataset, we use front-facing and backward-facing images. For the latter, we have inverted the order of the sequence. Additionally, we augment the training dataset by horizontal mirroring of images. Since the nuScenes dataset contains images from a left-driving country, the introduced images remain within the domain. We use the 700/150 scene split for training and validation, as recommended by nuScenes, so that the training subset contains 52060 and the validation subset -- 5569 sequences of length four. Images are resized from an initial 1600$\times$900 to 1024$\times$576 pixels and normalized to a range of [0,1]. For attack, we do not apply horizontally mirrored images, so that the training dataset contains 26030 images.

Since we focus on the detection of cars and pedestrians, the nuScenes classes \textit{adult, child, construction worker, police officer, stroller} and \textit{wheelchair} are  consolidated into one new class \textit{pedestrian}. As a second class we define \textit{car}, which corresponds to the nuScenes class \textit{vehicle.car}. We do not include other vehicles available in the nuScenes labels (e.g. busses, motorcycles, trucks, and trailers) to minimize intra-class variance.

 \begin{figure*}[t!]
	\includegraphics[width=1.0\textwidth]{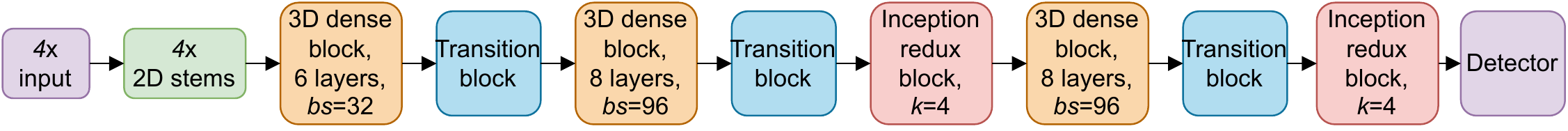}
	\caption{Architecture of the temporal feature network used for baselines. \textit{bs} denotes the bottleneck size. Inception block uses $k×\times1\times1$ convolutions for depth reduction}
	\label{fig:network}
\end{figure*}

\subsection{Baselines}

The baselines follow the late slow fusion architecture (see Figure \ref{fig:network}) proposed by Weber et al.~\cite{weber2021}. The input is a sequence of images with temporal distance $\Delta t = 500ms$, whereas prediction is learned for the last input frame. Each input image is processed by a separate 2D convolutional stem, being identical to the 2D DSOD detector~\cite{Shen_2017_ICCV}. Stems are followed by a series of 3D dense blocks, intervened with transition blocks. The subsequent Inception redux blocks aim at reducing the temporal depth by half. Finally, a detector subnet similar to that of YOLOv2~\cite{redmon2017yolo9000} follows.

For each dataset, we deliberately defined new anchor boxes to have high IoU scores with the bounding boxes in the training data. For this, we analyzed all objects in the training dataset and clustered them using k-means with IoU as a distance metric. As a result, we obtained eight anchor boxes, which comprise various box forms to detect both pedestrians (upright vertical rectangles) and cars (horizontal rectangles).

We train the baselines on an NVIDIA RTX 2080 Ti GPU. KITTI models are trained for 50 epochs, whereas nuScenes models are trained for 100 epochs. For validation, the PASCAL VOC implementation of mAP is used. The mAP is always calculated for a confidence threshold of $0.01$ and nms threshold of $0.5$.

\begin{table}[t]
     \begin{center}
 		\begin{tabular}{|r | c | c | c | c | }
 			\hline
 			\textbf{Model / Attack} & \textbf{No attack} & \textbf{Universal} & \textbf{Universal} & \textbf{Universal}  \\
 			& & \textbf{patch} & \textbf{noise} & \textbf{noise,} \\ 
 			& &  & \textbf{$\epsilon=5/255$} & \textbf{$\epsilon=10/255$} \\ 
 			\hline \hline
 			Baseline &  74.82  &  13.62 & 34.73  &  2.08   \\ 
 			5-PGD AT with patch &  66.47  & 52.18 & 24.46 & 23.07  \\
 			5-PGD AT with noise, $\epsilon=5/255$ &  62.77  &  42.82 & 61.52 &  61.75 \\
 			5-PGD AT with noise, $\epsilon=10/255$ &  53.24  & 32.15  & 53.24 &  53.07 \\
 			\hline
 		\end{tabular}
 \end{center}
\caption{$AP_{car}$ in \% of the 1-class KITTI baseline and AT models}
\label{tab:kitti_oneclass}
\end{table}

\begin{table}[t]
     \begin{center}
 		\begin{tabular}{|r | c | c | c | c | }
 			\hline
 			\textbf{Model / Attack} & \textbf{No attack} & \textbf{Universal} & \textbf{Universal} & \textbf{Universal}  \\
 			& & \textbf{patch} & \textbf{noise} & \textbf{noise,} \\ 
 			& &  & \textbf{$\epsilon=5/255$} & \textbf{$\epsilon=10/255$} \\ 
 			\hline \hline
 			Baseline &  50.93 &  4.36  & 34.49 & 4.43   \\ 
 			5-PGD AT with patch &  50.86  & 44.18  & 23.88 & 14.83  \\
 			5-PGD AT with noise, $\epsilon=5/255$ &  41.95  &  27.90  & 41.95 &  41.35 \\
 			5-PGD AT with noise, $\epsilon=10/255$ &  38.50  & 25.76 & 38.49 &  38.49 \\
 			\hline
 		\end{tabular}
 \end{center}
\caption{$mAP$ in \% of the 2-class KITTI baseline and AT models}
\label{tab:kitti-twoclass}
\end{table}

We train models to detect either only objects of the class \textit{car} (1-class models) or objects of the classes \textit{car} and \textit{pedestrian} (2-class model). Tables \ref{tab:kitti_oneclass} and \ref{tab:kitti-twoclass} show performance of the 1-class and 2-class KITTI baselines. Whereas average precision for the class \textit{car} is comparable for both models ($AP_{car}$ reaches 74.82\%  for the 1-class model vs. 73.21\% for the 2-class), the worse performance for the underrepresented class \textit{pedestrian} (28.65\%) explains the overall worse mAP of the 2-class model.

Tables \ref{tab:nusc-oneclass} and \ref{tab:nusc-twoclass} summarize results on the nuScenes baselines. Due to class imbalance (108K annotated cars vs. 48K annotated pedestrians), the results for the 2-class baseline for the underrepresented class \textit{pedestrian} are again significantly worse. 

The results achieved on the KITTI and nuScenes baselines, described below, are comparable if the opposite is not stated.

\begin{table}[t]
     \begin{center}
 		\begin{tabular}{|r | c | c | c | c | }
 			\hline
 			\textbf{Model / Attack} & \textbf{No attack} & \textbf{Universal} & \textbf{Universal} & \textbf{Universal}  \\
 				& & \textbf{patch} & \textbf{noise} & \textbf{noise,} \\ 
 			& &  & \textbf{$\epsilon=5/255$} & \textbf{$\epsilon=10/255$} \\ 
 			\hline \hline
 			Baseline &  73.20 & 6.20 & 9.03 & 0.15  \\ 
 			5-PGD AT with patch & 74.04   & 61.89 & 27.71 & 27.65   \\
 			5-PGD AT with noise, $\epsilon=5/255$ & 72.24  & 3.51 & 71.85 & 68.19  \\
 			\hline
 		\end{tabular}
 \end{center}
\caption{$AP_{car}$ in \% of the 1-class nuScenes baseline and AT models}
\label{tab:nusc-oneclass}
\end{table}

\begin{table}[t]
     \begin{center}
 		\begin{tabular}{|r | c | c | c | c | }
 			\hline
 			\textbf{Model / Attack} & \textbf{No attack} & \textbf{Universal} & \textbf{Universal} & \textbf{Universal}  \\
 			& & \textbf{patch} & \textbf{noise} & \textbf{noise,} \\ 
 			& &  & \textbf{$\epsilon=5/255$} & \textbf{$\epsilon=10/255$} \\ 
 			\hline \hline
 			Baseline &  27.55 & 0.98 & 0.74 & 0.16  \\ 
 			5-PGD AT with patch &  28.69  & 17.95 & 7.02 & 0.53   \\
 			5-PGD AT with noise, $\epsilon=5/255$ & 27.10 & 12.83 & 25.78 & 20.69   \\ \hline
 			AT with reused patches &  27.24  & 9.13 & 6.01 & 0.48  \\
 			R-FGSM AT with noise, $\epsilon=5/255$ & 27.51 & 2.72 & 0.24 & 0.02   \\
 			\hline
 		\end{tabular}
 \end{center}
\caption{$mAP$ in \% of the 2-class nuScenes baseline and AT models. For this model, AT with reused patches and R-FGSM AT were additionally evaluated}
\label{tab:nusc-twoclass}
\end{table}

\subsection{Adversarial Noise Attack}

\begin{figure}[t]
\centering
\begin{subfigure}[t]{0.485\linewidth}
    \includegraphics[width=1.0\textwidth]{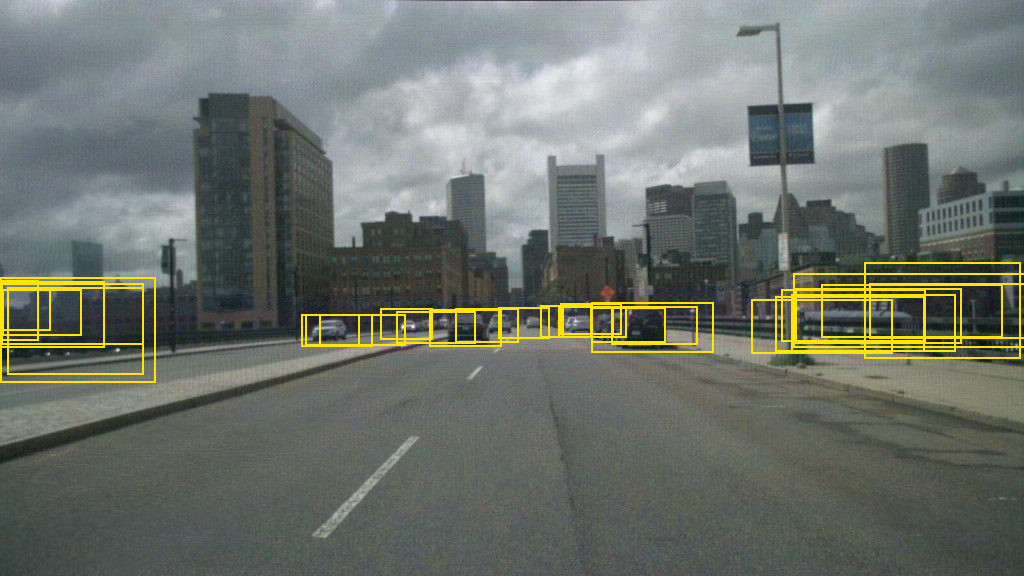}
  	\caption{Untargeted attack with gradient ascent}
  	\label{bad_noise_prediction}
\end{subfigure}
\begin{subfigure}[t]{0.485\linewidth}
  \includegraphics[width=1.0\textwidth]{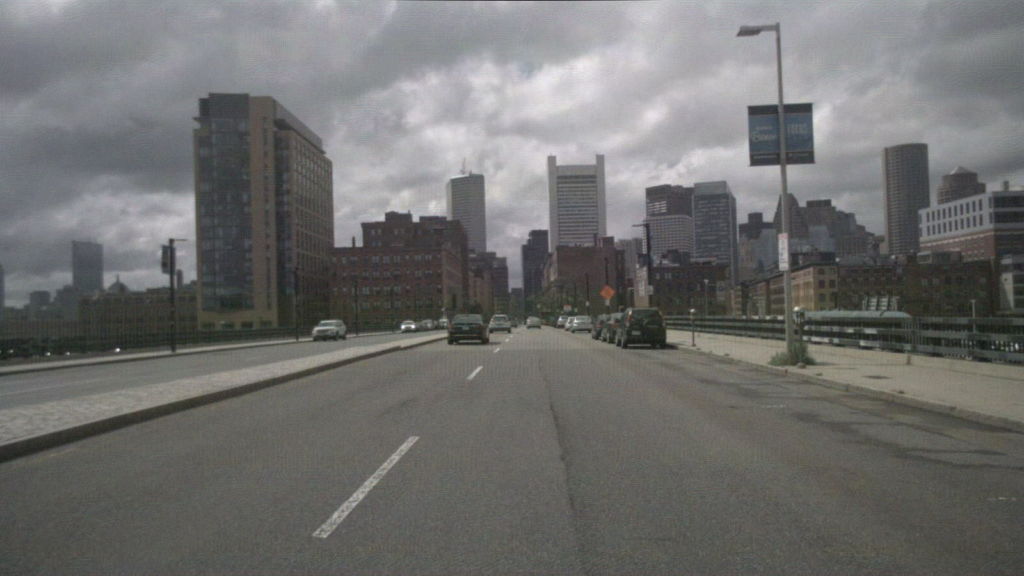}
  	\caption{Targeted attack for an empty label with gradient descent}
 	\label{noise_pred}
\end{subfigure}
\caption{Predictions of the 2-class nuScenes baseline on images attacked with universal noise, $\epsilon=5/255$}
\label{fig:soft-importance}
\end{figure}

For a universal noise attack, we have initially applied the proposed universal PGD with the changes motivated above. However, it turned out, that this attack leads to the detection of nonexistent new objects (see Figure~\ref{bad_noise_prediction}). The loss maximization goal apparently favors creating new features that resemble objects rather than preventing the detection of existing objects. We have therefore adapted the attack algorithm by replacing the gradient ascent with the gradient descent on an empty label. Figure~\ref{noise_pred} shows, that adversarial noise generated using targeted PGD on an empty label can successfully suppress all objects present in the input. All perturbations are trained using the Adam optimizer for 100 epochs.

\subsection{Adversarial Patch Attack}

To generate a universal patch, we apply PGD with unsigned gradients using the Adam optimizer for 100 epochs. We evaluated patches of size $71\times71$, $51\times51$ and $31\times31$. As expected, the largest patch led to a stronger attack and was used in the following experiments. Note, that a $71\times71$ patch still takes only about 1\% of the image area both in the case of KITTI and nuScenes images.

\begin{figure*}[h]
\centering
\begin{subfigure}[t]{0.485\linewidth}
    \includegraphics[width=1.0\textwidth]{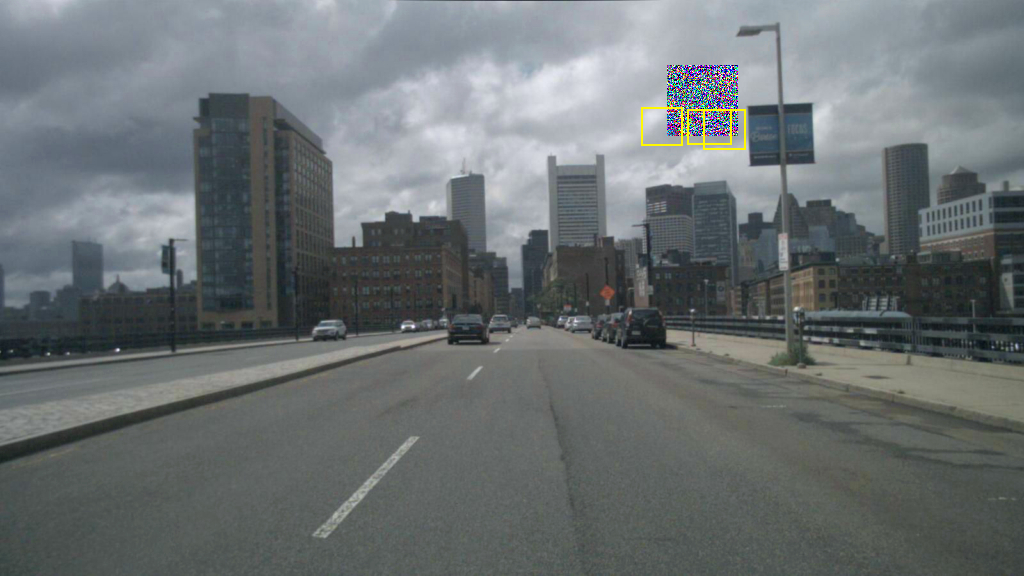}
  	\caption{Untargeted attack with gradient ascent on the 2-class model}
\end{subfigure}
\begin{subfigure}[t]{0.485\linewidth}
  \includegraphics[width=1.0\textwidth]{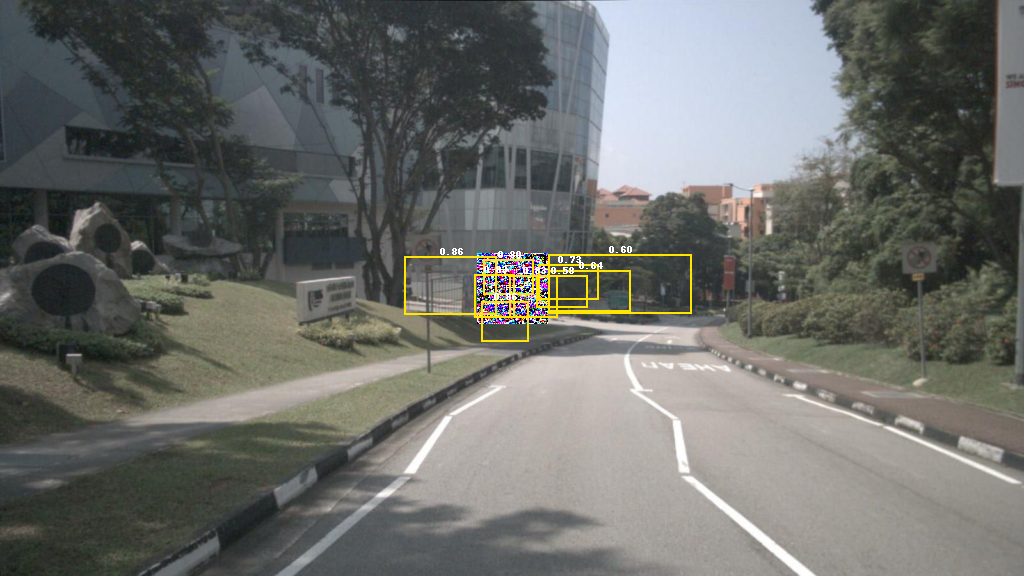}
  	\caption{Untargeted attack with gradient ascent on the 1-class model}
\end{subfigure}
\caption{Predictions of the nuScenes baselines on images attacked with universal patch}
\label{fig:patch-attacks}
\end{figure*}

To evaluate the impact of patch position, we evaluated a total of 33 patch positions. Average precision for different positions fluctuated only within few percentage points, so patch position apparently has only a minor impact on its attack strength. 

\subsection{Adversarial Training}

We apply adversarial training as a method to improve the robustness of the studied models. We consider K-PGD AT and two additional AT strategies: (1) reusing the patches, already generated for the baseline, during the training and (2) R-FGSM.

In the case of K-PGD AT, creating an adversarial example iteratively with $k$ steps increases the number of forward and backward propagations by a factor of $k$. We have therefore used a small $k = 5$ per adversarial attack during training and drastically increased the learning rate of the Adam optimizer to make the attacks possible. Training 5-PGD AT both on nuScenes and KITTI thus took five times longer than the corresponding baseline, both for the adversarial noise and the adversarial patch. 

In the case of the pre-generated patches for adversarial training, we first generated a pool of patches against the baseline as described above. We then trained a model, while adding a randomly chosen patch at each training step with a 50\% probability. The AT with reused patches involved no generation of new patches, therefore its duration is comparable to regular model training. 

Finally, R-FGSM AT was trained with adversarial noise with $\epsilon=5/255$.

\section{Evaluation}

\subsection{Attacks on 1-class vs. on 2-class Baselines}
 
\begin{figure}[h!]
\centering
 \begin{subfigure}[t]{0.485\linewidth}
	\includegraphics[width=\textwidth]{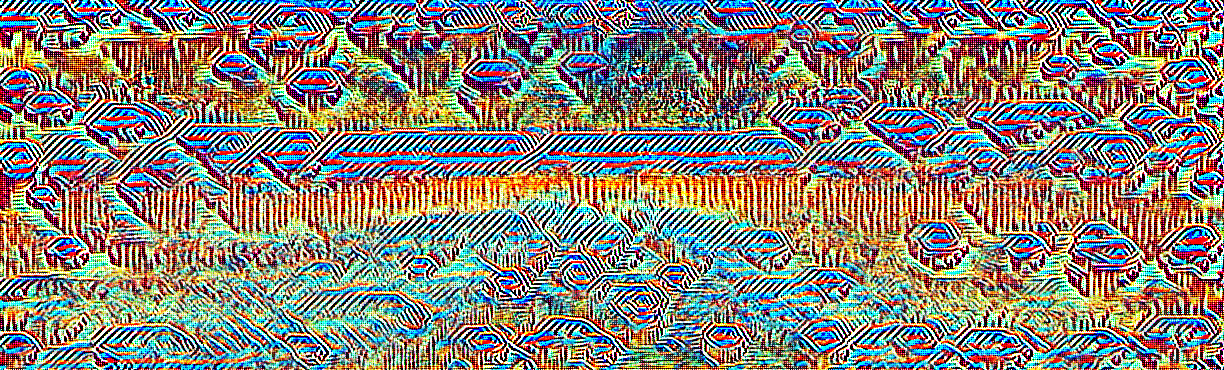}
  	\caption{KITTI 1-class baseline}
  \end{subfigure}
 \begin{subfigure}[t]{0.485\linewidth}
  	\includegraphics[width=\textwidth]{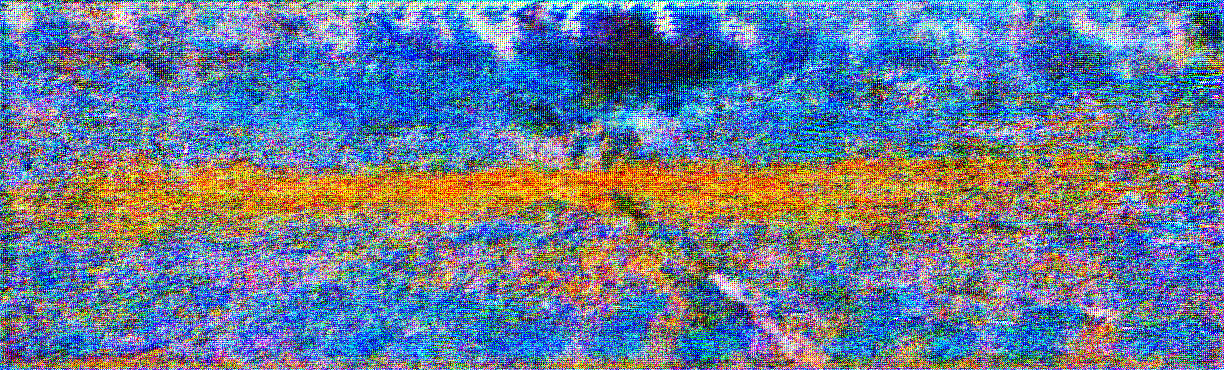}
  	\caption{KITTI 2-class baseline}
 \end{subfigure}
 
    \begin{subfigure}[t]{0.485\linewidth}
  	\includegraphics[width=\textwidth]{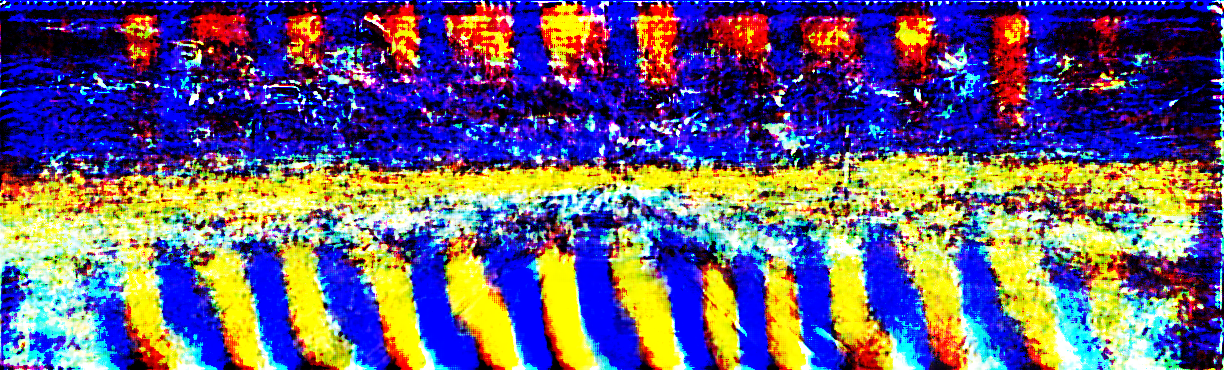}
  	\caption{KITTI 1-class robustified}
 \end{subfigure}
  \begin{subfigure}[t]{0.485\linewidth}
  	\includegraphics[width=\textwidth]{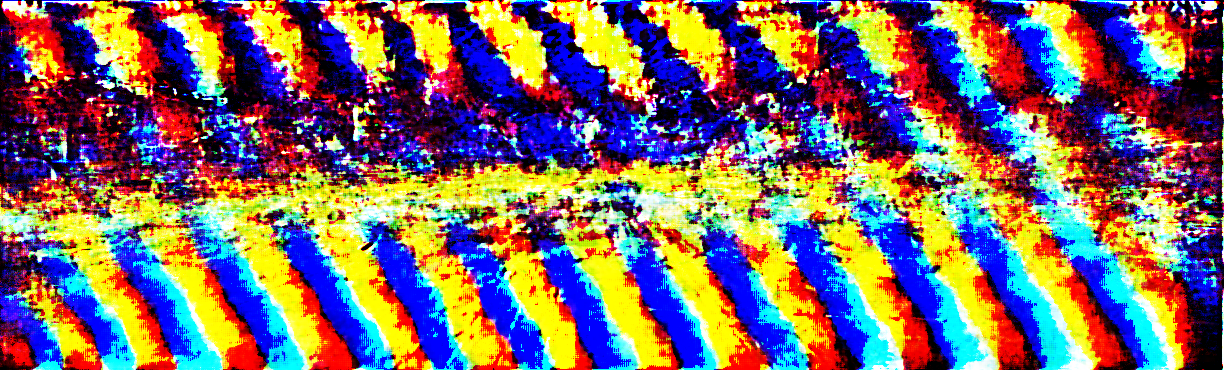}
  	\caption{KITTI 2-class robustified}
 \end{subfigure}

  \begin{subfigure}[t]{0.485\linewidth}
  	\includegraphics[width=\textwidth]{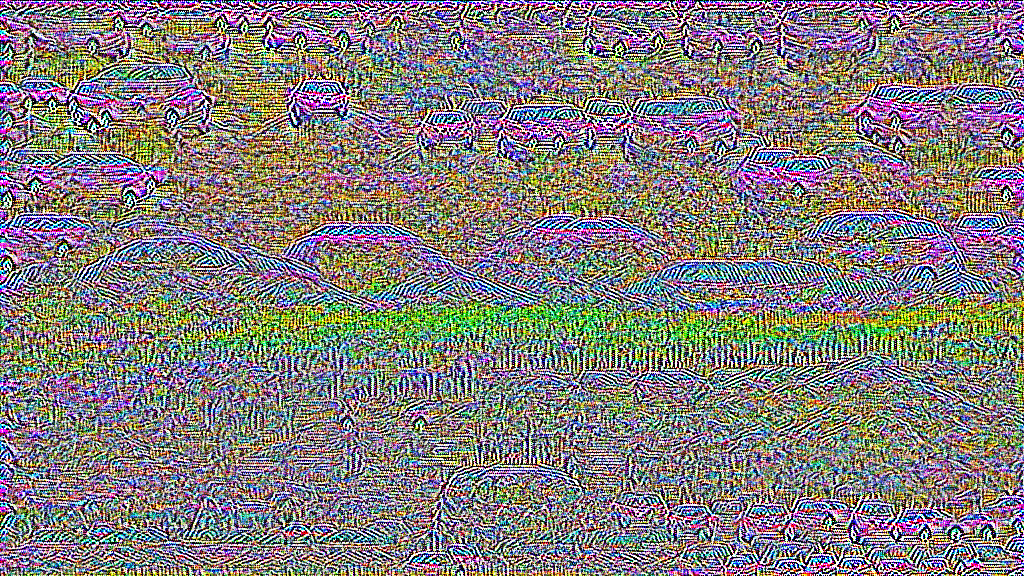}
  	\caption{nuScenes 1-class baseline}
 \end{subfigure}
  \begin{subfigure}[t]{0.485\linewidth}
  	\includegraphics[width=\textwidth]{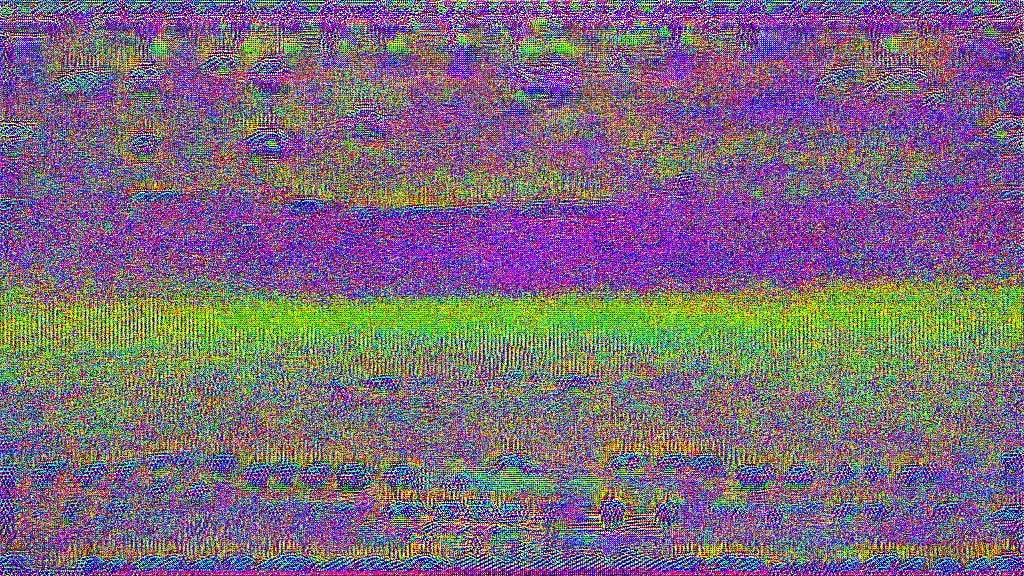}
  	\caption{nuScenes 2-class baseline}
 \end{subfigure}

  \begin{subfigure}[t]{0.485\linewidth}
  	\includegraphics[width=\textwidth]{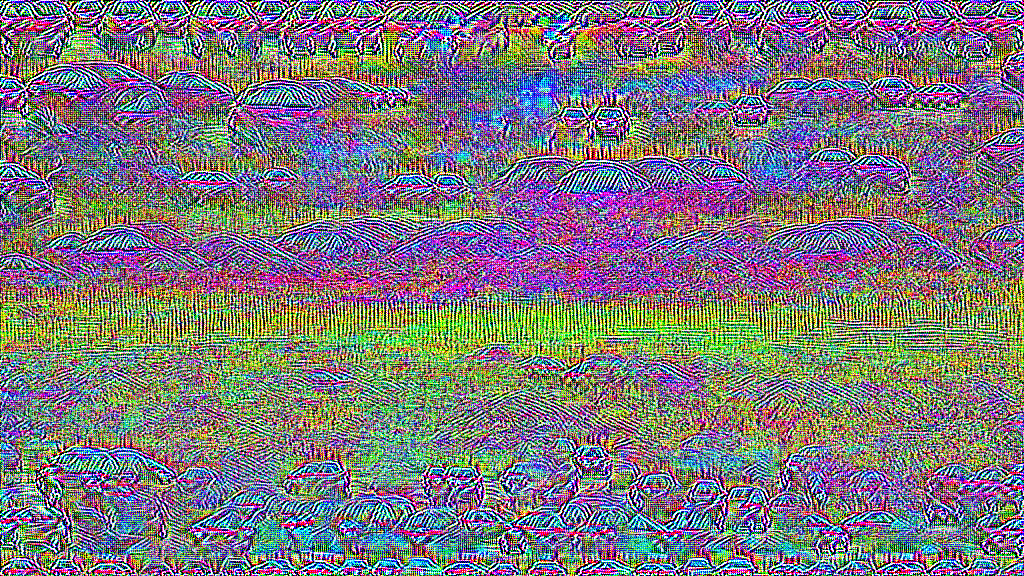}
  	\caption{nuScenes 1-class robustified}
 \end{subfigure}
  \begin{subfigure}[t]{0.485\linewidth}
  	\includegraphics[width=\textwidth]{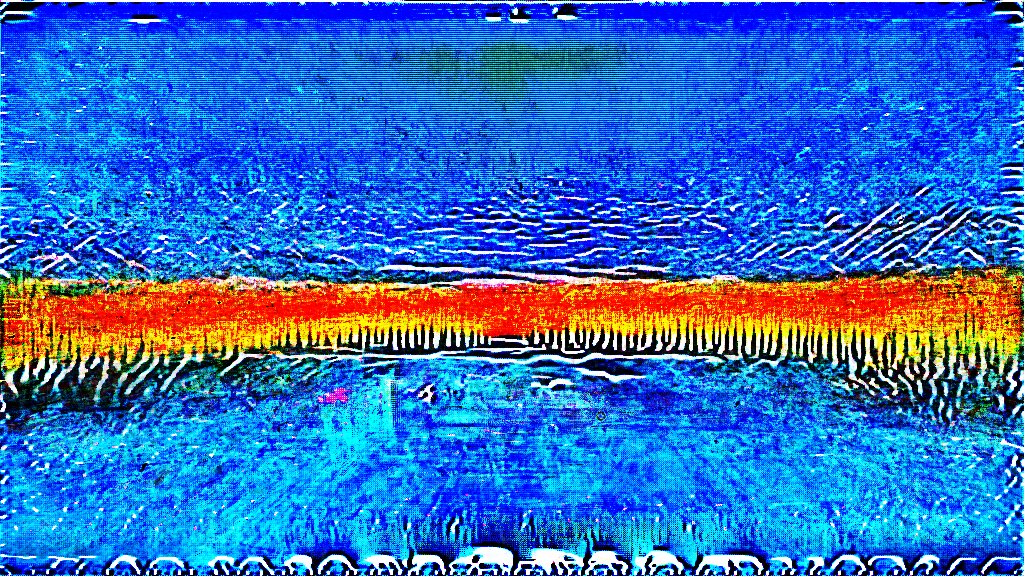}
  	\caption{nuScenes 2-class robustified}
 \end{subfigure}

 \caption{Universal noise with $\epsilon=5/255$, generated to attack baselines and models, robustified via 5-PGD AT with noise, $\epsilon=5/255$. Original pixel value range [-5,5] mapped to [0,255] for better visibility}
 \label{fig:universal-noise-1class2class}
 \end{figure}

Visual assessment of the adversarial noise patterns helps to understand, how the attack functions. We observed different behavior of attacks on 1-class and 2-class models. In particular, universal noise, generated to attack 1-class baselines evidently contains structures resembling cars, whereas noise attacking 2-class models exhibits no such patterns (see Figure \ref{fig:universal-noise-1class2class}). Apparently, the attack aims at mimicking existing objects, if they all belong to one class.

Adversarial patches, generated for both types of models, however, look similar. Patch-based attacks on nuScenes baselines tend to detect non-existing cars in a patch (see Figure \ref{fig:patch-attacks}), whereas attacks against the KITTI 2-class baseline rather find pedestrians in a patch. This might be explained by a different portion of pedestrians in the corresponding datasets. Patches against nuScenes 2-class model never mimicked pedestrians, because they are highly underrepresented in the training data.

 \subsection{Impact of Temporal Horizon}

Temporal fusion models have better performance due to the incorporation of the temporal history. To assess which portion of the history is enough to attack the model, we perform the evaluation exemplary with the $\epsilon=10/255$ adversarial noise attack on the 1-class KITTI baseline. We have attacked single frames using adversarial noise, which was initially generated for the whole input sequence of length four and with adversarial noise, generated for the corresponding portion of the input (see Table \ref{tab:kitti-part}). We observed, that perturbations, deliberately generated for specific frames, work better when attacking them, than those generated for the whole input sequence. For both cases, the attack works the best, when frames, immediately preceding the current frame, are attacked. On contrary, attacking only the oldest frame leads to the worst results. Also, perturbing only the frame for which the prediction is done and not attacking the temporal history at all leads to a significantly weaker attack. Finally, the more preceding temporal history frames are attacked, the better the results. 

These results confirm, that the later images in the input sequence are more important for the prediction. Furthermore, single attacked images that appear later in the input sequence, cause larger error than those which appear earlier. 

Furthermore, we evaluate the impact of the temporal horizon. In addition to the already evaluated models with four input images, we also evaluated models with a smaller sequence length. In particular, for the 1-class nuScenes model, we observe about 10\% for each reduction of the number of input images: 73.20\% mAP for the temporal history of length four, 63.12\% for the length three and 52.18\% for the length two. The attack strength also decreases correspondingly. We thus conclude that a larger temporal horizon helps to enhance not only the performance on the clean data, but also the adversarial robustness.

\begin{table}[t]
\centering
\centering
 		\begin{tabular}{|c |  c | c |}
 			\hline 
 			Attacked sequence part & Noise generated  & Noise generated  \\
 			$t_{-3}$ \ \ \ \  \ \ $t_{-2}$ \ \ \ \  \ \ $t_{-1}$ \ \ \ \ \ \  $t$ & for all
four inputs &  for each evaluated case \\ \hline
 			\raisebox{-0.5\height}{\includegraphics[width=0.4\textwidth]{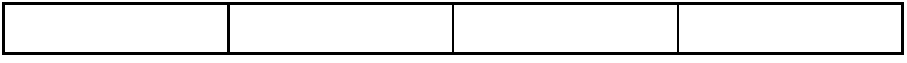}}   &74.82 &  74.82 \\ 
 			\raisebox{-0.5\height}{\includegraphics[width=0.4\textwidth]{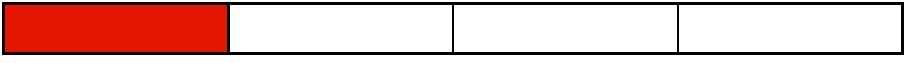}}   &70.68 &  58.46   \\
 			\raisebox{-0.5\height}{\includegraphics[width=0.4\textwidth]{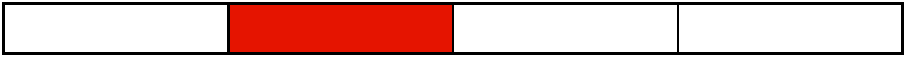}}   & 67.02 & 45.43\\
 			\raisebox{-0.5\height}{\includegraphics[width=0.4\textwidth]{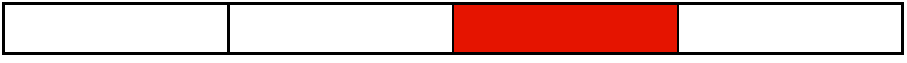}}   & 53.14 & 36.21 \\
 			\raisebox{-0.5\height}{\includegraphics[width=0.4\textwidth]{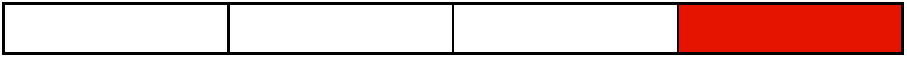}}   &23.91 &  5.77\\
 			\raisebox{-0.5\height}{\includegraphics[width=0.4\textwidth]{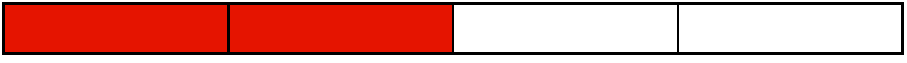}} & 50.84   &   22.14 \\
 			\raisebox{-0.5\height}{\includegraphics[width=0.4\textwidth]{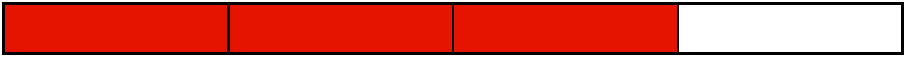}}   & 32.45 &  10.13    \\
 			\raisebox{-0.5\height}{\includegraphics[width=0.4\textwidth]{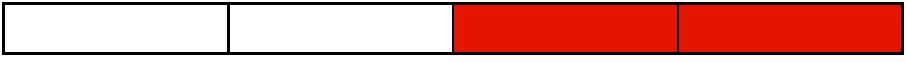}}   & 6.19 & 2.96 \\
 			\raisebox{-0.5\height}{\includegraphics[width=0.4\textwidth]{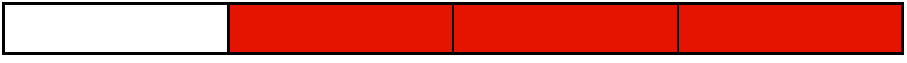}}   &3.80&  2.74 \\
 			\raisebox{-0.5\height}{\includegraphics[width=0.4\textwidth]{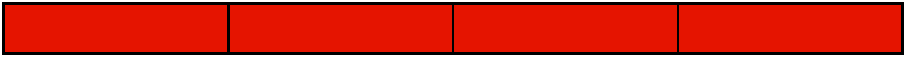}}   & 2.08  & 2.08 \\ \hline

 		\end{tabular}
\caption{Attacking a part of the sequence. The attacked frames are highlighted red, prediction is performed for the frame $t$. $mAP$ in \% is reported for the 1-class KITTI baseline, attacked with $\epsilon=10/255$ adversarial noise, generated either for all four inputs or for each evaluated case}
\label{tab:kitti-part}
\end{table}

\begin{figure}[t]
\centering
 \begin{subfigure}[t]{0.2\linewidth}
	\includegraphics[width=\textwidth]{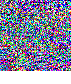}
  	\caption{Baseline}
  \end{subfigure}
 \begin{subfigure}[t]{0.2\linewidth}
  	\includegraphics[width=\textwidth]{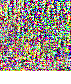}
  	\caption{AT with reused patches}
 \end{subfigure}
  \begin{subfigure}[t]{0.2\linewidth}
  	\includegraphics[width=\textwidth]{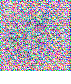}
  	\caption{5-PGD AT with patch}
 \end{subfigure}
  \begin{subfigure}[t]{0.2\linewidth}
  	\includegraphics[width=\textwidth]{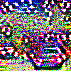}
  	\caption{5-PGD AT with noise}
 \end{subfigure}
 \caption{Universal patches to attack different nuScenes 2-class models}
 \label{fig:patches}
 \end{figure}

\subsection{Robustness of the Adversarially Trained Models}

To evaluate the robustness of the robustified models, we attack them again with newly generated universal patches and noises. Tables \ref{tab:kitti_oneclass}-\ref{tab:nusc-twoclass} demonstrate the results.

All models robustified via AT demonstrate performance similar to the baseline on non-attacked data and almost no accuracy drop on the malicious data for the nuScenes models and a small drop for the KITTI models. For the KITTI dataset, we have additionally evaluated AT with adversarial noise with $\epsilon=5/255$ and $\epsilon=10/255$ (see Tables \ref{tab:kitti_oneclass} and \ref{tab:kitti-twoclass}). The larger epsilon leads to worse performance on clean data due to larger perturbation.

Moreover, we have evaluated AT with reused patches and R-FGSM AT on the 2-class nuScenes model (see Table \ref{tab:nusc-twoclass}). As expected, the defended model with reused patches is less robust to attacks than the one which was trained with 5-PGD. Surprisingly, the R-FGSM AT method has completely failed to defend against attacks. We explain this behavior with the \textit{catastrophic overfitting} phenomenon, mentioned in the original work by Wong et al. ~\cite{wong2020fast} and in a more recent study by Andriushchenko et al.~\cite{andriushchenko2020understanding}, which challenges the original claim that using randomized initialization prevents this overfitting.

Figure~\ref{fig:patches} compares patches, generated for the adversarially trained models with the patch generated against the nuScenes 2-class baseline. In the case of patch reuse, the patch contains more green and yellow pixels than the original patch. In the case of K-PGD adversarial training, the patch is brighter and contains more white pixels. Interestingly, the patch generated for a model, which was adversarially trained with adversarial noise, is the only one that contains structures, resembling a car.

Figure~\ref{fig:universal-noise-1class2class} compares universal noises, generated to attack the KITTI and nuScenes baselines and the corresponding model defended via 5-PGD AT with $\epsilon=5/255$ noise. Both  contain a streak of color at the horizon line and wave-like patterns at the bottom. We again conclude, that perturbation attacking robustified models exhibit more complex structure.

\subsection{Robustness of the AT-trained Models against Per-instance Attacks}
Finally, we examine whether adversarially trained models also become robust against per-instance attacks. For this, we take an exemplary input sequence and generate adversarial perturbations against it. Each attack is trained for 1000 steps.

\begin{figure}[t]
\centering
 \begin{subfigure}[t]{0.485\linewidth}
	\includegraphics[width=\textwidth]{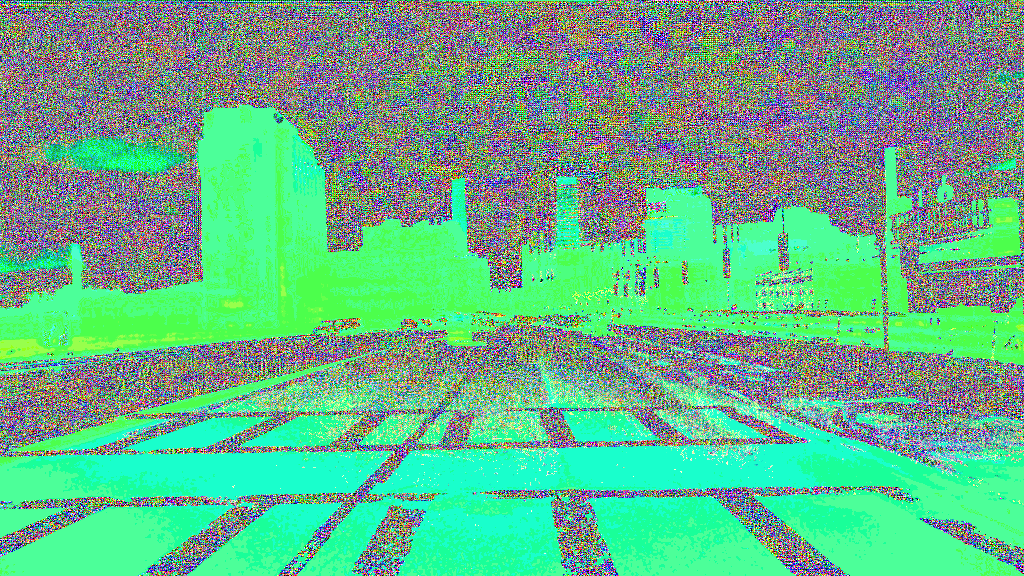}
  	\caption{Baseline}
  \end{subfigure}
  \begin{subfigure}[t]{0.485\linewidth}
  	\includegraphics[width=\textwidth]{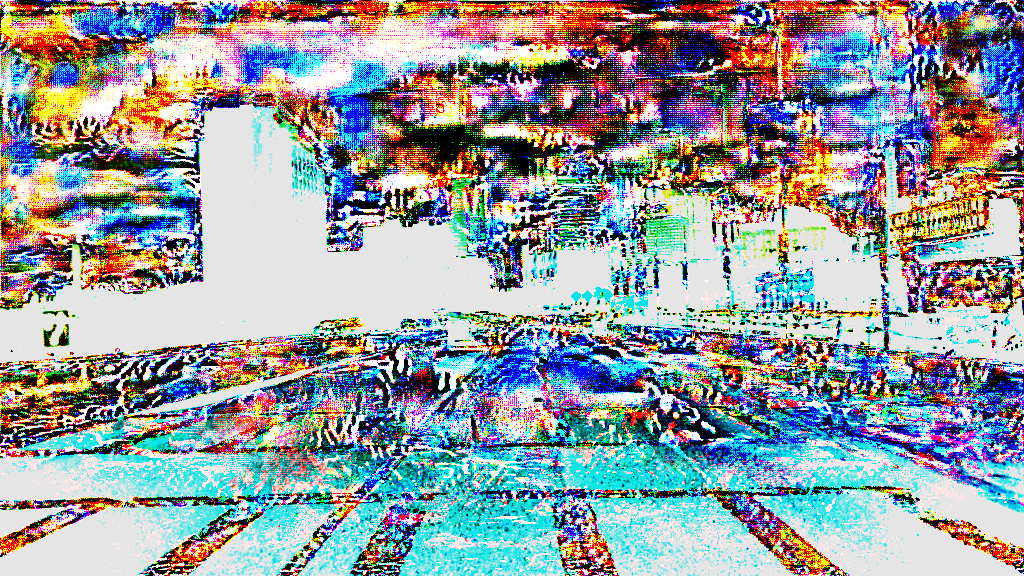}
  	\caption{5-PGD AT with noise}
 \end{subfigure}
 \caption{Per-instance adversarial noise with $\epsilon=5/255$ generated to attack nuScenes 2-class model on a single input sequence. Original pixel value range [-5,5] mapped to [0,255] for better visibility}
 \label{fig:perinstance-noise}
\end{figure}

\begin{figure}[t]
\centering
 \begin{subfigure}[t]{0.2\linewidth}
	\includegraphics[width=\textwidth]{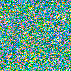}
  	\caption{Baseline}
  \end{subfigure}
  \begin{subfigure}[t]{0.2\linewidth}
  	\includegraphics[width=\textwidth]{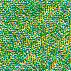}
  	\caption{5-PGD AT with patch}
 \end{subfigure}
  \begin{subfigure}[t]{0.2\linewidth}
  	\includegraphics[width=\textwidth]{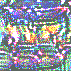}
  	\caption{5-PGD AT with noise}
 \end{subfigure}
 \caption{Per-instance adversarial patch generated to attack nuScenes 2-class model on single input sequence}
 \label{fig:perinstance-patch}
\end{figure}

Per-instance noise attack (see Figure \ref{fig:perinstance-noise})  manage to completely suppress all detections of the corresponding model. Analogously to universal attacks, non-universal noise attacks against the hardened model look much more complex. Similarly, per-instance patches (see Figure \ref{fig:perinstance-patch}) against the robustified model show complex structures resembling cars. Interestingly, this patch is also unable to efficiently attack the model, several cars are still correctly detected after applying this patch.

Overall, while models hardened with the evaluated adversarial training strategies are very successful in resisting universal attacks while preserving high accuracy, they are still unprotected against per-instance attacks. Universal attacks are, however, much more feasible with regard to real-life settings.

\section{Conclusion}
In this work, we have studied the adversarial vulnerability of temporal feature networks for object detection. The architecture proposed by Weber et al.~\cite{weber2021} was used as an exemplary model under attack.

Our experiments on KITTI and nuScenes datasets have demonstrated that the studied temporal fusion model is susceptible to both universal patch and noise attacks. Furthermore, we have explored different adversarial training strategies as a defense measure. Out of the three evaluated methods, the 5-PGD approach with a per-instance adversarial noise has proven to be the most powerful. The R-FGSM strategy, however, has failed to defend against the studied attacks. 5-PGD adversarial training was able to withstand newly created universal attacks. The robustified networks have also demonstrated only a slight drop in performance on clean data. 

Our experiments with attacking a portion of the temporal history have demonstrated, that the frames, immediately preceding the current frame, have a greater impact on the model decision and thus lead to stronger attacks when manipulated. We have further observed, that reducing the temporal horizon leads to worse performance and adversarial robustness of the model.

We have compared the universal and per-instance perturbations generated to attack the baseline and the robustified models. In all cases, we observed that in order to attack a hardened neural network, the adversarial perturbation has to exhibit a much more complex structure. In particular, a universal patch against the most robust 5-PGD with noise contains a pattern resembling a car. 

Our adversarially trained models, however, remain vulnerable to non-universal attacks like per-instance-generated noise or patch. This stresses the need for further research in this area.

Since the computation time for adversarial training is still a bottleneck, adapting gradient re-usage strategies like~\cite{shafahi2019adversarial} or~\cite{zhang2019you} for models, which use different loss functions to learn an adversarial perturbation and to update the model weights, might be a promising line of research for future.

\subsubsection*{Acknowledgement}

The research leading to these results is funded by the German Federal Ministry for Economic Affairs and Climate Action within the project “KI Absicherung“ (grant 19A19005W) and by KASTEL Security Research Labs. The authors would like to thank the consortium for the successful cooperation.

\clearpage
%
%
\bibliographystyle{splncs04}
\bibliography{references}
\end{document}